\documentclass[conference]{IEEEtran}
\IEEEoverridecommandlockouts
\usepackage{cite}
\usepackage{amsmath,amssymb,amsfonts}
\usepackage{algorithmic}
\usepackage{graphicx}
\usepackage{textcomp}
\usepackage{multirow}
\usepackage{xcolor}
\newcommand{\z}{\pmb{z}}

\usepackage{hyperref}
\def\BibTeX{{\rm B\kern-.05em{\sc i\kern-.025em b}\kern-.08em
    T\kern-.1667em\lower.7ex\hbox{E}\kern-.125emX}}
\begin{document}

\title{A Self-supervised Framework for Improved Data-Driven Monitoring of Stress via Multi-modal Passive Sensing}

% \author{\IEEEauthorblockN{Anonymous}}
\author{\IEEEauthorblockN{Shayan Fazeli\IEEEauthorrefmark{1}, Lionel Levine\IEEEauthorrefmark{1}, Mehrab Beikzadeh\IEEEauthorrefmark{1}, Baharan Mirzasoleiman\IEEEauthorrefmark{1}, \\ Bita Zadeh\IEEEauthorrefmark{2}, Tara Peris\IEEEauthorrefmark{3}, Majid Sarrafzadeh\IEEEauthorrefmark{1}} \\\IEEEauthorblockA{\IEEEauthorrefmark{1}\textit{Computer Science Department, UCLA, Los Angeles, US} \\
\IEEEauthorrefmark{2}\textit{Department of Psychology, Chapman University, Los Angeles, US} \\
\IEEEauthorrefmark{3}\textit{The Jane and Terry Semel Institute for Neuroscience and Human Behavior, UCLA, Los Angeles, US} \\
} \\
$\texttt{shayan@cs.ucla.edu, lionel@cs.ucla.edu, mehrabbeikzadeh@cs.ucla.edu, baharan@cs.ucla.edu}$, \\ $\texttt{bzadeh@chapman.edu, tperis@mednet.ucla.edu, majid@cs.ucla.edu}$
}

\maketitle

\begin{abstract}
Recent advances in remote health monitoring systems have significantly benefited patients and played a crucial role in improving their quality of life. However, while physiological health-focused solutions have demonstrated increasing success and maturity, mental health-focused applications have seen comparatively limited success in spite of the fact that stress and anxiety disorders are among the most common issues people deal with in their daily lives.  In the hopes of furthering progress in this domain through the development of a more robust analytic framework for the measurement of indicators of mental health, we propose a multi-modal semi-supervised framework for tracking physiological precursors of  the stress response. Our methodology enables utilizing multi-modal data of differing domains and resolutions from wearable devices and leveraging them to map short-term episodes to semantically efficient embeddings for a given task. Additionally, we leverage an inter-modality contrastive objective, with the advantages of rendering our framework both modular and scalable. The focus on optimizing both local and global aspects of our embeddings via a hierarchical structure renders transferring knowledge and compatibility with other devices easier to achieve. In our pipeline, a task-specific pooling based on an attention mechanism, which estimates the contribution of each modality on an instance level, computes the final embeddings for observations. This additionally provides a thorough diagnostic insight into the data characteristics and highlights the importance of signals in the broader view of predicting episodes annotated per mental health status. We perform training experiments using a corpus of real-world data on perceived stress, and our results demonstrate the efficacy of the proposed approach in performance improvements\footnote{
% Codes will be released.
Codes are available at \href{https://github.com/shayanfazeli/tabluence}{https://github.com/shayanfazeli/tabluence}
}.
\end{abstract}


% \shayan{paper type: 6 pages + 1 extra = short based on \href{https://ieeeichi.github.io/ICHI2023/call_for_papers.html}{this page}}
\section{Introduction}
% - the importance of mental health in general
% ---------------------------------------------
The rising epidemic of mental health disorders,  worsened by the recent COVID-19 pandemic, speaks to the growing need for effective and timely management of mental health disorders. 

The pandemic led to an increase in the need for mental health services, while concurrently, given the circumstances surrounding the outbreak, limited access to traditional modalities of care. This necessitated the explosion in the usage of alternative mechanisms to deliver mental health services, mainly through remote formats  \cite{tsamakis2020covid}.

For that reason, in spite of increased barriers to access, unprecedented levels of funding have  gone into programs to address mental health issues among the general public. 
%% government spending
For instance, in 2020 alone, the United States government spent around $280$ billion dollars on mental health services \cite{thewhitehouse_2022}.

% how therapists deal with it and treatments etc.
% ----------------------------------------
Therefore, even as the pandemic, and associated restrictions on in-person activities, have subsided, the gaps it revealed in traditional in-person-based therapeutic services persist, and demand for remote solutions remains high. 

% importance of remote monitoring of it
% ----------------------------------------
While much of the focus of remote mental health services has been around the use of video conferencing, instant messaging, and other modes of communication to facilitate interactions between therapists and patients, the use of mHealth technology and passive monitoring has the potential to be equally impactful at addressing barriers to care and gaps in monitoring.
Furthermore, by leveraging personal digital devices equipped with numerous sensors that are capable of monitoring many aspects of an individual's physiology and lifestyle (e.g., heart rate, activity level), remote health monitoring provides a novel pathway to not only monitor existing indicators of mental health, but also to improve upon our understanding mental health disorders and their impacts on one's life.

% importance of anxiety and a summary of its effects
% ----------------------------------------
Stress, commonly defined as "physical, mental, or emotional strain or tension," is a widespread problem with numerous potential causes.
According to the American Institute of Stress, $73\%$ of people suffer from acute bouts of stress to a degree of magnitude that impacts their mental well-being.
Incidents of Anxiety often manifest similarly to stress, however, it is notable that it is not always immediately tied to a specific triggering or inciting event and may take longer to resolve. 

All told, both stress and anxiety problems are very common, to the extent that most adults have been affected by at least one anxiety-related disorder \cite{any_anxiety,apa_stress2020}.
Anxiety-related disorders can have a significant negative impact on the quality of life, leading to other mental health disorders such as depression, as well as causing physical health problems \cite{mayo2018_anxiety}.

In contemplating improved means to address mental health challenges generally, and anxiety-related disorders specifically, it is notable that a critical part of modern healthcare involves accurate and efficient tracking of individuals' well-being through time.
Examples include tracking athletes and their training trajectories, and patients' rehabilitation exercises \cite{seshadri2019wearabled,gwak2019extra,valente2022multi,gwak2022internet,zhao2020design}.
Compared to physiological health, the mental health domain is less investigated in the context of remote health monitoring. 
This is largely due to a confluence of reasons.
For one, the statistical sufficiency of observations obtained via data-driven approaches is not often intuitively clear (e.g., can one draw a conclusion regarding depression from the number of phone calls?).
Another reason is that the data required for enabling the use of artificial intelligence (AI) is often not readily available or exclusive due to privacy and regulatory concerns. 

The works in this domain, therefore, have mostly focused on longer-term patient phenotyping (e.g., classifying patients into bipolar disorder vs healthy) \cite{hassantabar2022mhdeep}.

While these high-level labels are useful, they can be limited in their utility, as stress often manifests as an emotional and physiological response of an individual to a triggering event, and can occur to anyone regardless of a formal diagnosis. 
For example, arguing with someone and being anxious about a deadline are instances of interpersonal and work-related stress that are liable to occur to anyone regardless of the existence of a pre-existing mental health disorder.
Furthermore, the highly localized and temporary nature of these shorter-term episodes, which given the scarcity of data, makes making the most out of the available observations critical.

% what we propose and our contributions
% ----------------------------------------
Inspired by the advancements in the domain of self-supervised learning, we propose a multi-modal self-supervised learning framework to learn the context of stress response from continuous physiological readings.
This proposed setup addresses the following challenges and concerns  regarding data-driven monitoring of stress and anxiety:
\begin{itemize}
    \item The proposed method is inherently modular with regards to the different modalities of data, and therefore proper data-layer transforms allows leveraging various devices (e.g., smartwatches and wearable sensors different from ours) to learn efficient representations for  health monitoring.

    \item The self-supervised component allows training the network with a higher level of granularity and makes training more efficient.
    This is especially needed as the amount of labeled data available is often limited and costly to acquire, in contrast to sensor data that is generally trivially available in large quantities.

    \item The use of the attention mechanism enables a diagnostic view of the system, allowing the researchers to look into the empirical connection between various modes of data for specific monitoring tasks, counteracting the masking effect of many deep-learning frameworks on interpretability.

    \item In developing this framework, we conducted experiments on real-world data collected on perceived stress and have shown that this approach improves the performance compared to prior work leveraging early-fused embeddings of the same benchmark dataset.
\end{itemize}

\section{Related Works}
% anxiety and stress
% ----------------------------------------
Stress and anxiety-related disorders are common mental health challenges.
Such disorders can have significant negative impacts on people's lives, including higher chances of depression and suicide as well as associated comorbidities with physical health issues \cite{mayo2018_anxiety}.
Unfortunately, in many cases, these issues  remain inadequately treated due to challenges ranging from lack of viable access to therapeutic services to associated stigmas with utilization \cite{any_anxiety,apa_stress2020}.
However, even when an individual decides to seek psychotherapeutic help to alleviate these problems, challenges persist in the diagnosis and effective treatment of their disorder. 
At the inception of care, the steps to diagnose and monitor often include clinical evaluation and comparing personalized symptoms to standardized criteria, for example, the Diagnostic and Statistical Manual of Mental Disorders (DSM-5), which is commonly used for this matter.
Researchers continue to study and improve the practicality and accuracy of guidelines such as DSM-5\cite{batelaan2012mixed,zimmerman2017measuring}, but there are challenges in converting aggregated and generalized diagnostic criteria, down to episodic-level incidents of stress and anxiety.

% the disconnect between emotional and physiological stress
% ----------------------------------------
The most obvious approach to doing so leverages biometric data, extracted from wearable sensors embedded in smart devices, that measure a physiological stress response. 
However, while such data is incredibly valuable, and notably, sensing devices have become increasingly sophisticated at monitoring physiological stress, the resulting analyses are incomplete at best. 
This owes to the fact that from the standpoint of straightforward correlative analytics, it is known that there is not a direct monotonic correlation between the emotional perception of stress an individual may feel and the manifestation of the underlying physiological stress response.
A meta-analysis in the social stress domains, for instance, has recently shown that merely $25\%$ of studies in the domain demonstrated a significant correlation between physiological stress and perceived emotional stress \cite{31}.

Given that self-reports of perceived stress often do not contain information on the physiological stress response, understanding the complex relationship between the two becomes a crucial matter\cite{31}.
It is also plausible to assume that such complexity also arises from various other confounding factors (e.g., demographics, occupation, and other mental health disorders such as attention-deficit hyperactivity disorder (ADHD) can influence how prone someone is to stress).

This discrepancy has meaningful impacts on the utility of passive detection of stress based largely on physiological indicators. 
While sensors may be returning accurate readings on physiological stress, if they do not align with the user's own perceptions of stress, notably if they fail to properly account for moments when a user feels acute emotional distress, then it will demotivate further engagement with a mental health platform. 

% health monitoring in general
% ----------------------------------------
This hindrance comes in spite of considerable progress that has been made in recent decades regarding the capabilities and efficacy of personal digital devices, including smartwatches, smartphones, and wearable devices.
This fact has made such devices attract a lot of research and commercial attention, employing them for various monitoring objectives \cite{15,16}.

These monitoring approaches focus primarily on fitness and health-related aspects, resulting in a large body of research and countless commercialized applications. Examples include tracking athletes' training, detecting falls for the elderly, tracking post-surgery therapeutic and rehabilitation exercises, and posture correction \cite{darabi2017heart,vilarinho2015combined,gwak2019extra,wile2014smart}.
% mental health monitoring
% ----------------------------------------
While the central focus of health monitoring applications has undoubtedly been on physical health, a wide range of research works has focused on understanding the relationship between observations obtained leveraging digital devices and some aspects of individuals' mental health status.
% importance of "passive" sensing and why it is more beneficial
It is noteworthy that a primary goal in designing smart and automated approaches for mental health monitoring has to do with proposing meaningful passive-sensing tools so that informative observations regarding health status can be made by eliminating or diminishing the need to interfere with users' daily activities or request repeated active interactions.

As a remote mental health monitoring task, social anxiety was studied in the previous literature, and it was shown that analyzing trajectories obtained via smartphone location services can paint a comprehensive picture concerning individuals' proneness to it.
To do so, the movements and the nature of locations visited (which were obtained by cross-referencing location data with a map API) were taken into consideration, and the hypothesis of whether or not such corpus is informative for recognizing the presence of social anxiety was tested \cite{boukhechba2018predicting,chow2017using,huang2016assessing,gong2019understanding}.
Smartphones have also been helpful in developing an understanding of anxiety \cite{levine2020anxiety}.

Another choice of hardware for gathering data pertinent to health data is application-specific wearable sensors. For instance, wearable electrocardiogram (ECG) sensors were used to recognize perceived anxiety via pattern recognition \cite{king2018predicting}.

% imporance of smartwatch (a section before moving on to the related work on smartwatch, which is mainly mhdeep and our earlier work)
Smartwatches have a unique position amongst the wide range of various commonly used digital devices.
They are in close contact with the skin and, given their attachment user's wrist, which is a distal point of a major appendage, make it possible to obtain most measurements (e.g., activity) at higher accuracy, as well as enabling additional measurements such as heart-rate or pulse oximeter.
In case of the need for brief questions, interactions, or Ecological Momentary Assessments (EMAs), smartwatches can also be used to issue messages and acquire responses and entries by the user \cite{15,16,17}.
Additionally, smartwatches are prevalent, and relying on them as the hardware for health applications provides a better alternative in most cases to application-specific wearable devices in terms of cost, comfort, and user-friendliness.
% mHealth anxiety works
% ----------------------------------------
Data-driven analyses leveraging smartwatches' sensory readings have been successful at the problem of patient classification for bipolar disorder, schizoaffective disorder, and depression \cite{hassantabar2022mhdeep}.
It has also been shown that physiological readings made by basic smartwatch sensors enable efficient modeling of perceived stress response \cite{fazeli2022passive}.

% semisupervised learning, and anxiety works such as mhdeep
% ----------------------------------------
In the health analytics domain, data and human annotations are often limited.
Therefore, dealing with overfitting and memorization is a crucial matter.
Additionally, it is beneficial to go beyond the limited number of human annotations available in training efficient inference pipelines.
Less reliance on annotations by focusing on unsupervised and self-supervised approaches has received a lot of research attention in recent years \cite{chen2020simple,caron2020unsupervised,chen2021exploring,grill2020bootstrap}.
The core idea in most works in this area is that comparing and contrasting  the latent representations of examples that are expected to share certain similarities (e.g., augmented versions of the same image) can benefit the trained weights and help with regularizing the learned  decision boundaries \cite{zhang2017mixup}.

% difference
% ----------------------------------------
In short, this work is primarily focused on addressing the limitations in the previous literature on remote mental health monitoring.
The previous works do not go beyond leveraging scarcely available annotations in training network parameters and mainly rely on data augmentation to improve their performance.
They do not focus on encapsulation in embedding different modalities, which can be an obstacle in employing optimal encoders for each modality and can hinder transfer learning.
Additionally, they do not focus on the interpretability of the inference pipeline, which is crucial in health-related applications.

\begin{figure*}
    \centering    
    \includegraphics[width=.8\textwidth]{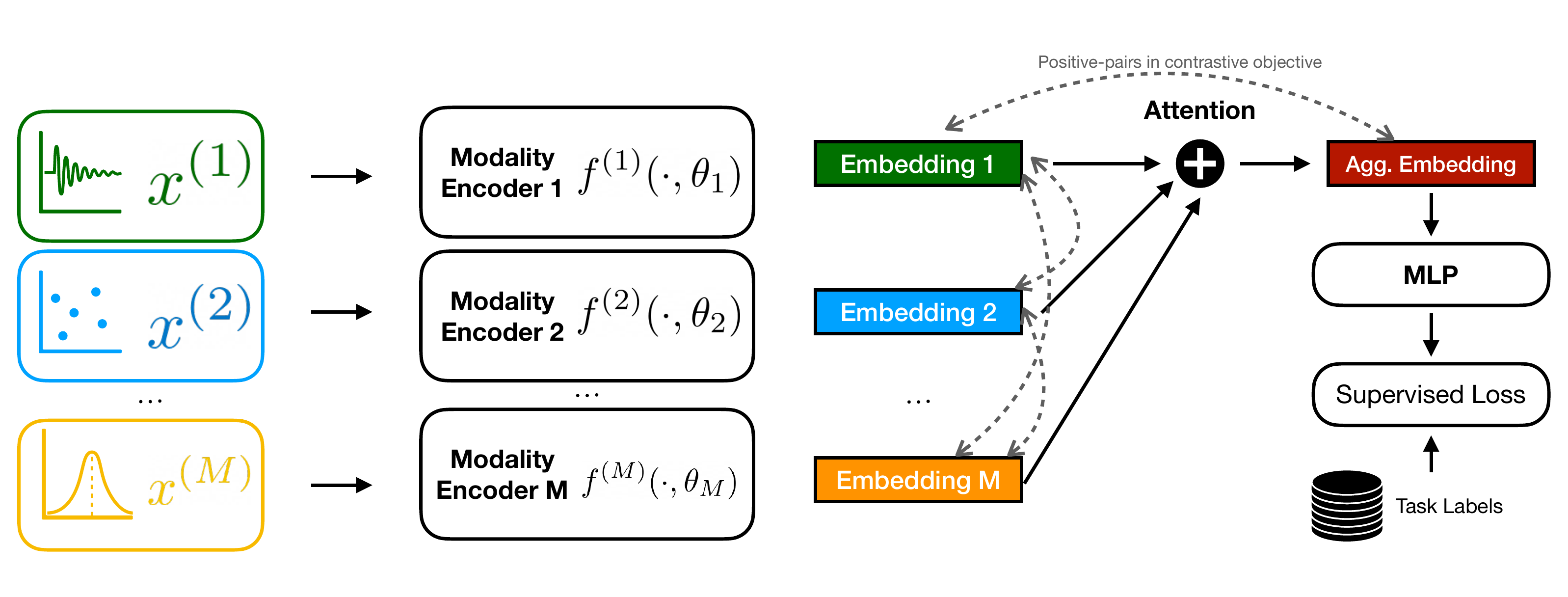}
    \caption{
    Our proposed multi-modal self-supervised learning pipeline.
    Modality-specific data from different distributions are encoded through dedicated encoders and mapped to a shared latent representation space. The aggregated embedding of the segment is then computed by applying attention pooling to the modality-specific representations.
    A self-supervised contrastive objective aligns this aggregate embedding and the mode-specific representations.}
    \label{fig:mmssl}
\end{figure*}

To address these challenges, this work proposes a framework for leveraging smartwatch-based sensor-driven data to recognize {\it perceived stress}, enabling a novel approach to remote mental health monitoring.
Our proposed inference pipeline is modular and hierarchical and is composed of modality-specific embedding branches.
The final embedding is computed via a task-specific attention-pooling mechanism, which also provides an interpretation of the estimated contribution of each modality's information to the last embedding.
During training, we leverage an inter-modality contrastive objective so as to encourage consistency among the predictions and tune all encoder branches.
Figure \ref{fig:mmssl} depicts the overview of our proposed framework.
The details of our approach are discussed in the next section.

\section{Methodology}

Consider a cohort of $P$ individuals undergoing a study wearing a smartwatch-based remote monitoring system.
This wearable setup allows for the collection of sensory readings pertinent to users' exhibited physiological and activity patterns throughout the day.
In our case, the study in question monitors the connection between readings made by the smartwatch, which are mostly related to physiological signals, health and activity status, and short-term {\it perceived} stress reported by the user.
The smartwatch's readings can thus be grouped into features corresponding to several modalities: $m \in \{1,2,\cdots,M\}$.
This grouping depends mainly on the nature of the features, as well as the setup and the interface provided by the smartwatch.
From each modality, we have a sequence of observed feature vectors:
\begin{equation}
\pmb{x}_m^{(p)} = \{x^{(p)}_{m,t}\}_{t \in [T^{(p)}_{m,\text{max}}]}
\end{equation}

From a user's timeline, we extract short-term timespans, each of which corresponds to an {\it episode} $e$, which is the result of filtering the timeline and restricting it to the episode's timespan: $e=(t_{\text{start}}, t_{\text{end}})$:
\begin{equation}
    \pmb{x}_{m,e}^{(p)} = \{x^{(p)}_{m,t} \in \mathbf{x}_m^{(p)} \mathopen|\mathclose t \in e \}
\end{equation}

We have a parameterized domain-specific\footnote{The term {\it domain} in this manuscript refers to the observation type, for example, a Transformer-based Language Model could efficiently represent data from textual domain, and there could be multiple {\it modalities} with their observations being text data, each represented by their own specific encoder.} encoder $f(\cdot; \theta_m)$ for each modality $m \in [M]$, which performs the task of mapping the observed data from this modality to a {\it shared} semantic space $\mathcal{S}$:
\begin{equation}
f(\cdot;\theta_m): \mathcal{X}_m \rightarrow \mathcal{S} \quad \forall m \in [M]
\end{equation}
Hence, the latent embedding denoted by $z^{(p)}_{m,e}$ can be found as follows:
\begin{equation}\label{eq:findinglatents}
z^{(p)}_{m,e} = f(\pmb{x}_{m,e}^{(p)};\theta_m) \quad \forall m \in [M]
\end{equation}

One could argue that it is plausible to assume that the contributions of observations from different modalities to the final prediction on a specific task follow a non-uniform distribution in most cases.
For instance, there is no reason to assume the statistical significance of heart-rate time series is the same as pulse oximeter readings for the task of stress detection.
Going one step further, such disparity can manifest itself in the level of {\it instance} representations as well.
To illustrate this further, consider a simple case of "missingness" in data or presence of noise.
This could mean that even though mode $m_1$, for example, is more informative (in expectation) to the task $\tau$, in an instance where the data from this group appears missing or clearly corrupt, the importance of other modalities could change respectively.
Therefore, we have designed a {\it modality importance} head, implemented as a fully connected pipeline, which determines the contribution of each mode by weighing their respective embedding vectors, which were projected to the same semantic space.
The first step to this attention-based pooling mechanism involves using the modality importance head and obtaining  (yet unnormalized) weight $a_i^{(m)}$ for the latent embedding of modality $m$'s information in an instance $i$:
\begin{equation}\label{eq:attn1}
    a_i^{(m)} \leftarrow g(\z_i^{(m)};\pmb{\psi}) \quad \forall m \in [M]
\end{equation}
This is followed by a softmax operation to make sure that the summation of the predicted contributions maps to unity, in other words, the contribution matrix is right stochastic:
\begin{equation}\label{eq:attn2}
\alpha_i^{(m)}=\frac{\exp(a_i^{(m)})}{\sum_{j\in[M]}\exp(a_i^{(j)})} 
\end{equation}
And thus the final aggregated latent is computed using these attention weights:
$z=\sum_{i=1}^M \alpha_i^{(m)} \cdot z_{m}$.

We leverage the cosine similarity $\phi(\cdot, \cdot)$ to measure the compatibility between the latent representation of each mode and the aggregate representation $z_i$.
\begin{equation}\label{eq:similarity}
\phi(\pmb{u},\pmb{v})=\frac{h(\pmb{u})^T\cdot h(\pmb{v})}{\|h(\pmb{u})\|_2\cdot \|h(\pmb{v})\|}
\end{equation}

In other words, we use the aggregated embedding $z$ as an anchor and define a contrastive objective to leverage the distances and inconsistency between the latent embeddings:
\vspace{-1mm}
\begin{equation}\label{eq:loss}
\mathcal{L}_{\text{cl}}=\frac{1}{|\mathcal{B}|}\sum_{i\in \mathcal{B}}\frac{1}{|\mathcal{M}|}\sum_{m\in \mathcal{M}}-\log\frac{\exp(\phi(\z_i^{(m)},\z_i)/\tau)}{\sum_{j\in \mathcal{B},j\neq i}\exp(\phi(\z_i^{(m)},\z_j)/\tau)} \vspace{-2mm}
\end{equation}

We have experimented with $\mathcal{L}_{\text{cl}}$ in the following training schemes:

\begin{itemize}
    \item {\it Pre-training}: Pre-training the model parameters by optimizing $\mathcal{L}_{\text{cl}}$ through a long training sequence.
    Afterward, start with the resulting weights as the initial point for the supervised fine-tuning of the model with the cross-entropy objective:
    \begin{equation}
    \mathcal{L}_{\text{cross-entropy}}=- \sum_{c\in\mathcal{C}} y_c \ln p_c
    \end{equation}
    In the equation above, $\mathcal{C}$ is the set of all classes (e.g., in our experiments, the two categories of stressed and non-stressed for each episode), and $p_c$ is the predicted probability of class $c$ for an observation, computed by passing representations through a final projection and Softmax layer.
    \item {\it Regularization}: Use $\lambda_{\text{reg}} \cdot \mathcal{L}_{\text{cl}}$ as a regularization term in the overall loss, and train the model by optimizing this loss simultaneously as the supervised learning objective.
\end{itemize}
There are several points worth remarking upon with regard to the comparison of these two training schemes. 
To begin with, deciding whether pre-training is going to lead to better generalization performance versus the regularization-based approach depends on model complexity, availability of data, and the challenges of the specific task that one is targeting.
That being said, the regularization approach is expected to be considerably faster than the two-stage pre-training and fine-tuning method, and in our experiments on the task of predicting stress labels, it led to better test performance as well.
% late-fusion vs early fusion and information advantage
% ----------------------------------------
% plug-and-play for inter-smartwatch compatibilities
% ----------------------------------------
% attention module and the contrastive setup (item-level vs mode-level)
% ----------------------------------------
% "regularization" alternative for "quicker" and "better performance", cite NCR
% ----------------------------------------
\section{Experiments}
% a discussion on the experiments
\subsection{Data}
\begin{figure}[h!]
    \centering
    \includegraphics[width=0.48\textwidth]{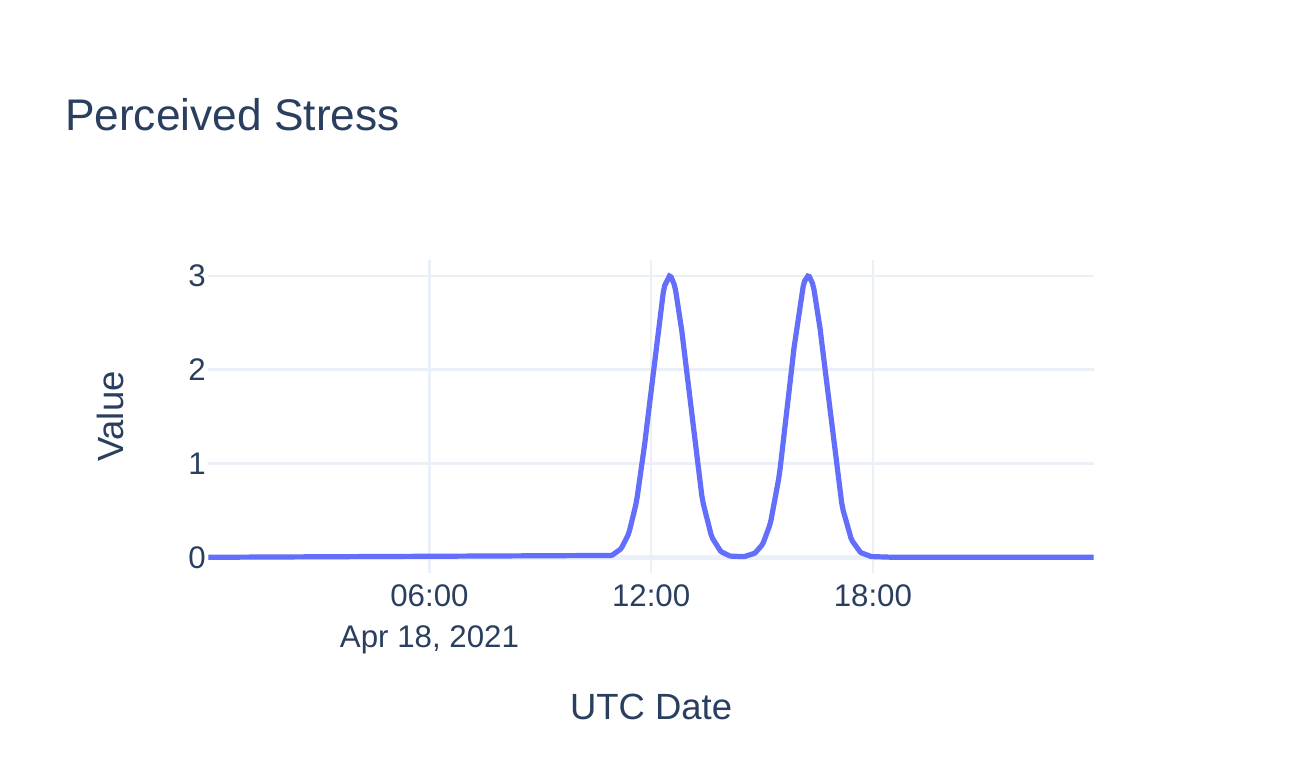}
    \caption{A sample portion of a continuous supervision signal generated based on user inputs, from which episode stress labels can be sampled}
    \label{fig:continuous_supervision_signals}
\end{figure}

\begin{figure}[h!]
    \centering
    \includegraphics[width=0.48\textwidth]{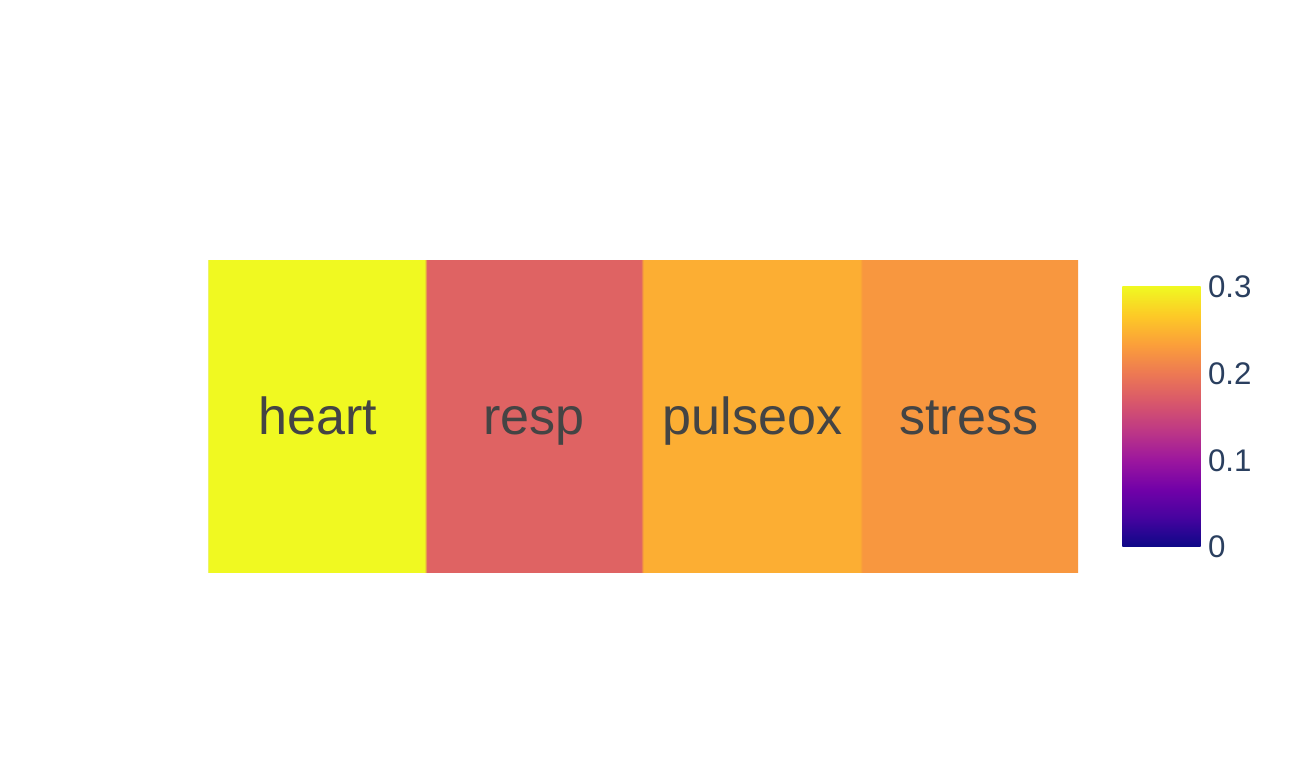}
    \caption{The average contribution of the four modalities to the final episode embeddings.}
    \label{fig:attn_interp}
\end{figure}

The cohort in this study consists of $14$ college students who are ex or active-duty members of the United States military.
The status of these individuals, both as military members as well as students, renders them an interesting cohort for our stress study, given that the individuals from both groups are known to be relatively more prone to experiencing stress.

\begin{table}[h]
\centering
\caption{Number of participants per category based on their duty status and service branch}
{\small
\label{tab:cohortcategories}
\resizebox{0.5\columnwidth}{!}{%
\begin{tabular}{l|l|}
\cline{2-2}
                                     & \textbf{Duty Status}    \\ \hline
\multicolumn{1}{|l|}{National Guard} & 2                          \\ \hline
\multicolumn{1}{|l|}{Active Duty}    & 3                          \\ \hline
\multicolumn{1}{|l|}{Veteran}        & 9                          \\ \hline
                                     & \textbf{Service Branch} \\ \hline
\multicolumn{1}{|l|}{Airforce}       & 2                          \\ \hline
\multicolumn{1}{|l|}{Marine}         & 3                          \\ \hline
\multicolumn{1}{|l|}{Navy}           & 4                          \\ \hline
\multicolumn{1}{|l|}{Army}           & 5                          \\ \hline
\end{tabular}%
}}
\end{table}

The smartwatch in this study was Garmin vivoactive 4S. Nevertheless, it is noteworthy that there is no component in the proposed methodology that limits the solution to this smartwatch.
The feature groups and various modalities in our configuration are shown in Table \ref{tab:modfocus}.
\begin{table}[]
\centering
\caption{Data modalities and the features they are focused on}
\label{tab:modfocus}
\begin{tabular}{|l|l|}
\hline
\textbf{Modality} & \textbf{Focus}     \\ \hline
Daily &
  \begin{tabular}[c]{@{}l@{}}heart-rate readings, \\ number of floors \\ climbed, \\ BMR kilocalories, \\ distance traveled,\\ activity levels, \\ aggregated HRV measures\end{tabular} \\ \hline
Pulse Ox          & SPO2               \\ \hline
Respiration       & Respiration rate   \\ \hline
Stress            & HRV-based readings \\ \hline
\end{tabular}
\end{table}

\subsection{Labeling}
The focus of this study has been on making predictions on {\it perceived} stress, for which the participants agreed to indicate the episodes in which they felt stressed and provide us with the intensity and timespans of these episodes.

For each record input to our system by an individual, we created a softened (via a Gaussian function) time-series per the following steps:
\begin{itemize}
    \item The peak (corresponding to the {\it mean} of this Gaussian function) is set to the given timestamp, or the midpoint of the timespan ($(t_{\text{start}} + t_{\text{end}})/2$).
    \item The standard deviation of $30$ minutes (scaled proportional to the length of time-span, if a time span of over one hour is provided).
    \item The magnitude of the peak point corresponds to the indicated  for the episode: $\{0,1,2,3\}$ for $\{\texttt {None}, \texttt{Low},\texttt{Medium}, \texttt{High}\}$, respectively.
\end{itemize}
The summation of these Gaussian signals comprises the  signal used as the primary supervision objective.
The way the labels are computed is by looking at the end-point of each episode, and its {\it stress} label is marked \texttt{True} if the value of this signal on that point is larger than a threshold of $0.5$, and \texttt{False} otherwise.
% \shayan{should we remove the subsections in the experiments section?}
\subsection{Modeling}
% separate rnn branches etc. description of the model
Our inference model is composed of a specific encoder for each modality.
In our case, each encoder is defined based on an initial mapping and normalization (via fully connected layers) followed by a bi-directional recurrent neural network (RNN) in long short-term memory (LSTM) configuration.
Specifically, the data from each modality was first projected to a $32$-dimensional vector via a multi-layer perception (MLP) with one hidden layer.
The output was then forwarded to the modality-specific bi-LSTM with the hidden-layer neuron count of $64$. The last stage for representing each modality was another fully-connected projection layer, generating a $32$-dimensional vector per modality, which were used as modality representations in our framework.

Note that the overall pipeline does not have any constraint on the local modality encoders as long as they share the final semantic space to which they project that modality's observations.
Given that we were mainly dealing with time-series data, we used RNNs to model each branch.
Nonetheless, modalities from substantially different domains and their encoders (e.g., Transformer-based Language Model for textual data) can also fit into the same system.

\subsection{Results}
Focusing on our real-world perceived stress corpus, we conducted experiments under the main settings of 1) supervised training baseline, 2) pre-training the contrastive objective and fine-tuning via supervised objective, and 3) training the supervised objective and simultaneously optimizing a scaled version of the contrastive term as a regularizing loss.
 
We observed that leveraging more features and following a late-fusion protocol for combining modality representations did lead to an improved generalization performance over the supervised setup proposed in \cite{fazeli2022passive}, which combined the features at the beginning of the pipeline. In the case of our cohort, training with contrastive regularization led to the best generalization on the unseen test data, and the results are shown in Table \ref{tab:modelresults}.
Note that, in general, it is hard to say which self-supervised setup (pre-training versus regularization) is best, as it could depend on other factors, including model complexity, optimization, data availability, and task difficulty.
That being said, our approach allows learning high-quality representations by optimizing the modality-contrastive objective via both of these setups.

Additionally, we focused on interpretability as well and leveraged the task-specific attention mechanism in our pipeline, which pools the representations from different modalities, to study the utility and contribution of observations from each feature group.
This enables the network to dynamically assign weights to each modality's latent representation (in the shared space) as it processes each instance, allowing us to study their contribution both per instance and in expectation for performing the desired task.
In Figure \ref{fig:attn_interp}, we have shown the results on this matter for the contrastive regularization setup\footnote{The label $\texttt{heart}$ in Figure \ref{fig:attn_interp} corresponds to the $\texttt{daily}$ modality's information, given that its main focus is heart-rate.}.
The results indicate that even though the contributions of the different modalities follow a non-uniform distribution as expected, none of them were ignored by the model and they all play a part in the final predictions.

\begin{table}[]
\centering
\caption{Performance comparison for the trained pipeline under different learning setups}
\label{tab:modelresults}
\begin{tabular}{|l|c|}
\hline
\textbf{Method}                                           & \multicolumn{1}{l|}{\textbf{Test Accuracy}} \\ \hline
Early-fusion + Supervised Training \cite{fazeli2022passive} & 64\% \\
Late-fusion + Supervised Training                                       & 66\%                                        \\
Late-fusion + Contrastive Pre-training + Fine-tuning                    & 70\%                                        \\
\textbf{Late-fusion + Supervised Training} &   \\
\textbf{+ Contrastive Regularization} & \textbf{73\%}

\\ \hline
\end{tabular}
\end{table}

\section{Discussion}
\subsection{Broader Impact}
% discussing transferrability and inter-smartwatch compatibilities etc. in the broader impact.
In the context of remote health monitoring, there are several factors addressing which is of paramount importance.
In what follows, we elaborate upon these factors and how the solution proposed in this work attempts to address them:
\begin{itemize}
\item{
{\it Affordability and Compatibility}: For the scalability of a proposed remote health monitoring framework, focusing on widely available devices that are sold at affordable price renders it easier to deploy the system.
In this work, we focused on basic physiological signals for which reading sensors are available in most commercially available smartwatches.
Nonetheless, the proposed methodology has no intrinsic limitation regarding the modalities used; thus, additional data available in often more expensive devices (e.g., galvanic skin response) can also be utilized in the same methodology, and the main requirement is providing a modality-specific encoder fit for the data domain.
Furthermore, this framework offers a more encapsulated view in representing different modalities as the observation from each can be embedded by a dedicated encoder first, and the contrastive objective encourages each local branch to optimize its parameters towards the given task as well.
This has clear advantages in terms of transferring knowledge as well, an example of which could be initializing each branch separately via pre-trained weights so as to prepare a better starting point for the model and optimization.
}
\item{
{\it Ease of use}: Optimizing a remote health monitoring with regards to minimizing the amount of required user interaction makes it easier for individuals to use the system.
This is why passive monitoring techniques are receiving more attention in the eHealth domain.
}
\item{
{\it Interpretation}: In all automated healthcare applications of machine learning, any insight and interpretation into what parts of the observation a model mostly focused on in determining the final decision, is crucial and can help experts better validate the system as well.
In this work, we incorporated a task-specific attention mechanism for pooling the representations from different modalities, which helps determine the weights assigned to each modality (per instance and in expectation) to perform the task efficiently.
}
\item{
{\it Limited Data}: The data availability for eHealth applications is often limited due to the difficulty and costs of conducting large-scale studies, the exclusivity of data, and privacy reasons.
It is, therefore, important to try to maximize the use of data in training inference pipelines.
This work combines label smoothing with inter-mode self-supervision objectives to go beyond self-reported supervision objectives.
}
\end{itemize}

\subsection{Limitations}
It is crucial to discuss the limitations of this work given the sensitive nature of dealing with health as its objective.
%% error in the data
In this work, we relied on self-reported entries to decide the supervision signals for individual timelines.
This has the issue of being prone to human error, as one might not accurately recall the time and extent to which one has felt stress.
Additionally, reports on the intensity of the felt stress are also subject to noise.
%% small data size and its challenges, use of ssl to address it to some extent
Another challenge is the small size of our dataset.
A primary reason behind our self-supervision component in this work was alleviating the negative impacts associated with the aforementioned limitations.

\subsection{Conclusion}
We proposed a remote health monitoring solution that is modular and multi-modal, thus, allowing the use of various encoders best suited for each modality. 
We proposed an instance-level attention mechanism to tune the contribution of each modality to the final representation and provide insight into the expected importance of each modality for the task at hand.
We conducted experiments with the proposed method to recognize perceived stress in short-term episodes and empirically demonstrated its superior performance over supervised training.

% - references
\bibliographystyle{IEEEtran}
\bibliography{main}

\end{document}